\documentclass[sigconf]{acmart}

\AtBeginDocument{%
  \providecommand\BibTeX{{%
    \normalfont B\kern-0.5em{\scshape i\kern-0.25em b}\kern-0.8em\TeX}}}

\copyrightyear{2021} 
\acmYear{2021} 
\setcopyright{rightsretained} 
\acmConference[SIGIR '21]{Proceedings of the 44th International ACM SIGIR Conference on Research and Development in Information Retrieval}{July 11--15, 2021}{Virtual Event, Canada}
\acmBooktitle{Proceedings of the 44th International ACM SIGIR Conference on Research and Development in Information Retrieval (SIGIR '21), July 11--15, 2021, Virtual Event, Canada}\acmDOI{10.1145/3404835.3463258}
\acmISBN{978-1-4503-8037-9/21/07}

\usepackage{tabularx}
\usepackage{multirow}
\usepackage[usestackEOL]{stackengine} 
\usepackage{cellspace}
\usepackage{graphicx}
\graphicspath{ {./images/} }
\usepackage{pifont}
\usepackage{balance}
\usepackage{enumitem}
\usepackage{makecell}

\newcommand{\hi}[1]{\textcolor{blue}{#1}}
\newcommand{\fa}[1]{\textcolor{black}{#1}}

\newcommand*\OK{\ding{51}}
\newcommand{\hjscalebox}{\scalebox{0.8}[0.8]}

\newcommand{\shrink}{\vspace*{-.9\baselineskip}}

\begin{document}

\fancyhead{} 

\title{Conversational Entity Linking: Problem Definition and Datasets}


\author{Hideaki Joko}
\affiliation{%
  \institution{Radboud University}
}
\email{hideaki.joko@ru.nl}

\author{Faegheh Hasibi}
\affiliation{%
  \institution{Radboud University}
}
\email{f.hasibi@cs.ru.nl}

\author{Krisztian Balog}
\affiliation{%
  \institution{University of Stavanger}
}
\email{krisztian.balog@uis.no}

\author{Arjen P. de Vries}
\affiliation{%
  \institution{Radboud University}
}
\email{a.devries@cs.ru.nl}


\begin{abstract}

Machine understanding of user utterances in conversational systems is of utmost importance for enabling engaging and meaningful conversations with users. Entity Linking (EL) is one of the means of text understanding, with proven efficacy for various downstream tasks in information retrieval. In this paper, we study entity linking for conversational systems.
To develop a better understanding of what EL in a conversational setting entails, we analyze a large number of dialogues from existing conversational datasets and annotate references to concepts, named entities, and personal entities using crowdsourcing. Based on the annotated dialogues, we identify the main characteristics of conversational entity linking.  Further, we report on the performance of traditional EL systems on our Conversational Entity Linking dataset, ConEL, and present an extension to these methods to better fit the conversational setting.
The resources released with this paper include annotated datasets, detailed descriptions of crowdsourcing setups, as well as the annotations produced by various EL systems. 
These new resources allow for an investigation of how the role of entities in conversations is different from that in documents or isolated short text utterances like queries and tweets, and complement existing conversational datasets.

\end{abstract}




\maketitle

\section{Introduction}
\label{sec:introduction}

Conversational systems are becoming increasingly important with the proliferation of personal assistants, such as Siri, Alexa, Cortana, and the Google Assistant. 
In this realm, understanding user utterances plays a crucial role in holding meaningful conversations with users---this process is handled by the natural language understanding (NLU) component in traditional task-oriented dialogue systems~\cite{Gao:2019:NAC}. 
A popular text understanding method, which has proven to be effective in various downstream tasks~\cite{Dalton:2014:EQF, Hasibi:2016:EEL, Xiong:2017:WED, Shang:2021:ERO, Hasibi:2017:DFS, Dargahi:2018:QUE}, is \emph{entity linking} (EL): the task of recognizing mentions of entities in text and identifying their corresponding entries in a knowledge graph~\cite{Balog:2018:EOS}. 
In this paper, we aim to investigate the role of entity linking in conversational systems.


Even though large-scale neural language models (like BERT~\cite{Devlin:2019:BER} and GPT-3~\cite{Brown:2020:LMF}) have repeatedly been shown to achieve high performance in various machine understanding tasks, 
these do not provide replacements for explicit auxiliary information from knowledge graphs.  Rather, the two should be seen as complementary efforts.  Indeed, augmenting neural language models with information from knowledge graphs has shown to be beneficial in a number of downstream tasks~\cite{Poerner:2020:EBERT, Peters:2019:KnowBERT, Zhang:2019:ERNIE}.  
Specifically, in the context of task-based conversational systems, NLU often relies on entities for fine-grained domain classification and intent determination. 
This makes EL for conversational systems even more important.


Despite its importance, research on EL for conversational systems has so far been limited. Traditional EL techniques that are used for documents~\citep{Balog:2018:EOS}, queries~\citep{Hasibi:2015:ELQ, Li:2020:ELQ}, or tweets~\citep{Meij:2012:ASM} are suboptimal for conversational systems for a number of reasons. 
First, unlike documents, conversation are not ``static,'' but are a result of human-machine interaction.  Here, the user can correct the system's interpretation of their request and the system can ask clarification questions to further its understanding of the user's intent. 
Second, conversations are informal and it is common in a conversation to make references to entities by their pronouns; e.g., ``my city,'' ``my guitar,'' and ``its population.''  A conversational system is expected to understand and handle such mentions of personal entities~\citep{Balog:2019:PKG}. 
Third, while entity linking in documents or queries tends to focus on proper noun entities~\cite{Carmel:2014:ERD, Hoffart:2011:RDN}, in a conversational setting all types of entities, including general concepts, can contribute to machine understanding of users' utterances.
In this paper, we aim to investigate these differences and develop resources to foster research in this area.


The main research question driving this research is the following: \emph{What does entity linking in conversations entail and how is it different from traditional entity linking?}  
To investigate this research question, we set out to analyze entity linking annotations for existing conversational datasets. We perform a thorough analysis of existing conversational datasets and select four of these for annotation.  These cover the three main categories of conversational problems~\cite{Gao:2019:NAC}: question answering (QA), task-oriented systems, and social chat (cf. Sect.~\ref{sec:dataset_section}).  We aim to annotate ``natural'' conversations, and therefore bias our selection of datasets towards those that are obtained using a Wizard-of-Oz setup.  Annotating dialogues, however, is an inherently complex task, where it is a challenge to keep the cognitive load sufficiently low for crowd workers.
This leads us to a secondary research question: \emph{What are effective designs for collecting large-scale annotations for conversational data?}  Although a large body of research exists on effective designs for collecting large-scale entity annotations~\cite{Adjali:2020:BME, Bhargava:2019:LMW, Finin:2010:ANE, Mayfield:2011:BCL}, to the best of our knowledge, there is no work on conversational data.  We run a number of pilot experiments using Amazon Mechanical Turk (MTurk) to select the best design and instruction. 
Based on these experiments, we develop the \emph{Conversational Entity Linking (ConEL)} dataset, consisting of of 100 annotated dialogues (708 user utterances) sampled from the QuAC~\cite{Choi:2018:QUA}, MultiWOZ~\cite{Zang:2020:MWO}, WoW~\cite{Dinan:2018:WWK}, and TREC CAsT 2020~\cite{Dalton:2020:CAT} datasets. 
To enable further study in conversational EL, we also annotate a separate sample of 25 WoW dialogues (containing references to personal entities) and all 25 manually rewritten dialogues of TREC CAsT 2020.


Our findings, obtained by analyzing the annotated dialougues, are as follows. 
\begin{itemize}[leftmargin=2em]
	\item Mentions of personal entities are mainly present in social chat conversations.
	\item While named entities are deemed to be important for text understanding, specific concepts are also found useful for understanding the intents of conversational user utterances.
	\item Traditional EL approaches fall short in providing high precision annotations for both concepts and named entities. This calls for a methodological departure for conversational entity linking, where concepts, named entities, and personal entities are taken into consideration.
\end{itemize}

\noindent
In summary, this work makes the following contributions:
\begin{itemize}[leftmargin=2em]
  \item To the best of our knowledge, ours is the first study on EL in conversational systems. We subdivide entities into three categories (named entities, concepts, and personal entities), and analyze the importance of each for conversational data. We further investigate different aspects of EL for three categories of conversational tasks: QA, task-oriented, and social chat.
  \item We investigate effective designs for collecting large-scale EL annotations for conversational data. 
  \item We make the annotated conversational datasets publicly available.\footnote{\url{https://github.com/informagi/conversational-entity-linking}} This data comes with detailed account of the procedure that was followed for collecting the annotations, which can be used for further extension of the collection. 
  \item As an additional (online) resource, we provide a comprehensive list of around 130 conversational datasets released by different research communities with a detailed comparison of their characteristics.  
\end{itemize}

\noindent
The resources provided in this paper allow for further investigation of entity linking in conversational settings, can be used for evaluation or training of conversational EL systems, and complement existing conversational datasets.
\section{Related Work}

The related work pertinent to this paper concerns entity linking in documents, queries, and conversational systems, as well as personal entity identification.

\subsection{Entity Linking}
\label{sec:entity_linking}
%
\emph{\textbf{Entity linking in documents.}}
Entity linking plays an important role in understanding what a document is about~\cite{Balog:2018:EOS}.
TagMe~\cite{Ferragina:2010:TAG} is one of the most popular EL tools, redesigned and improved by \citet{Piccinno:2014:FTM} and renamed to WAT.
Van Hulst et al.~\citep{vanHulst:2020:REL} presented REL,  which is an open source EL tool based on state-of-the-art NLP research.
Other state-of-the-art EL methods include DeepType~\cite{Raiman:2018:DTM}, Blink~\cite{Wu:2020:Blink}, and GENRE~\cite{Cao:2021:GENRE}.
Although these approaches are effective for documents, it is known that EL algorithms with high performance on general documents are less effective when applied to short informal texts like queries~\cite{Cornolti:2018:SMA}.

%
%
\emph{\textbf{Entity linking in queries.}}
Entity linking in queries poses new challenges due to the short and noisy text of queries, their limited context, and high efficiency requirement~\cite{Hasibi:2017:ELQ, Carmel:2014:ERD, Cornolti:2018:SMA}.
\citet{Cornolti:2018:SMA} tackled some of these challenges by ``piggybacking'' on a web search engine. Relying on external search engines, while being effective, hinders efficiency and sustainability of EL systems. 
\citet{Hasibi:2017:ELQ} studied this challenge with a special focus on striving a balance between effectiveness and efficiency.
These studies consider EL for a single query, while in conversational systems multiple consecutive user turns need to be annotated.


\emph{\textbf{Entity linking in conversations.}}
Research on conversational entity linking has been mainly focused on employing traditional entity linking and named entity recognition methods in conversational and QA systems~\cite{Kumar:2020:MIS, Chen:2016:CIM, Bowden:2018:SNE, Chen:2017:RWA, Vakulenko:2018:MSC, Li:2020:ELQ}. 
Entity linking is also used in multi-party conversations to connect mentions across different parts of dialogues and mapping to their corresponding character~\cite{Chen:2016:CIM}. This is a subtask of entity linking, referred to as character identification.
A close study to our work is~\cite{Bowden:2018:SNE}, where an entity linking tool, focused mainly on named entities,  is developed for open-domain chitchat.
In contrast to these works, we aim to understand EL for conversational systems, annotating conversations with concepts, named entities, and personal entities.

\subsection{Personal Entities}
\label{sec:personal_entities}
Dealing with the mention of personal entities (e.g., ``my guitar'') is important for  personalization of conversational systems.
Consider for example the user utterance ``Do you know how to fix my guitar?''
To answer this question, the system has to know more about the user's guitar type; e.g., ``Gibson Les Paul.'' This information may be available in the conversation history, previous conversations, or other sources (e.g., user's public information in social media). This information can be represented as RDF triples in the form of subject-predicate-object expressions $\langle e, p, e'\rangle$, e.g., \emph{$\langle$User, guitar, Gibson Les Paul$\rangle$}.
\citet{Li:2014:PKG} proposed a method to detect personal entities ($e$) and their corresponding predefined predicates ($p$) in conversations. Their approach consists of three steps: (1) identifying user utterances that are related to personal entities, (2) predicting entity mentions by classifying those utterances, and (3) finding the  personal entities.
\citet{Tigunova:2019:LBL} address the problem of identifying personal entities from implicit textual clues. They proposed a zero-shot learning method to overcome the lack of sufficient labelled training data~\cite{Tigunova:2020:CHA}.
All these studies focus on identifying predefined classes of predicates.
Extracting RDF triples without predefined relation classes has been studied in the context of open information extraction~\cite{Yates:2007:TRO, Fader:2011:IRO, Angeli:2015:LLS, Cui:2018:NOI}, but not in relation to personal entities.
In this study, we annotate conversations with personal entity mentions and their corresponding entities.

\section{Dataset Selection}
\label{sec:dataset_section}

There exists a large number of conversational datasets released by the natural language processing, machine learning, dialogue systems, and information retrieval communities. 
We made an extensive list of around 130 datasets,\footnote{This list is publicly available at:  \url{https://github.com/informagi/conversational-entity-linking}} extracted from ParlAI~\cite{Miller:2017:PAR} and other dataset comparison lists~\cite{Penha:2019:INT, Choi:2018:QUA, Hauff:2020:datasets}.
These datasets target three conversational problems~\cite{Gao:2019:NAC}:

\begin{itemize}[leftmargin=2em]
	\item \emph{Question answering (QA)}, where users ask natural language queries and the system provides answers based on a text collection or a large-scale knowledge repository.
	\item \emph{Task-oriented systems}, which assist users in completing a task, such as making a hotel reservation or booking movie tickets.
	\item \emph{Social chat}, where systems are meant to be AI companions to the users and hold human-like conversations with them.
\end{itemize}

\noindent
To obtain a comprehensive view of entity linking in conversational systems, we set out to analyze at least one dataset for each of the three main categories of conversational problems. 
To this end, we shortlisted datasets that resemble real conversations. That is, multi-domain and multi-turn  datasets, collected based on actual interactions between two humans. Datasets that are extracted from web services (e.g., Reddit and Stack Exchange) or created based on templates (e.g., bAbI~\cite{Sukhbaatar:2015:EEM}) were thus ignored.
To ensure that the selected datasets are sizable, they were required to contain at least 100 dialogues.
This list was further narrowed down by selecting relatively popular datasets based on citation counts and publication year.\footnote{While admittedly this is a loose measure, it helps to identify datasets that became widely accepted by the research community.}
By applying these criteria, nine datasets were shortlisted. 

In the final step, each dataset in our shortlist was closely examined, and at least one data set was selected for each conversational problem; see Table~\ref{tbl:extracted-candidates} for an overview of the selected datasets. 
The reasoning behind our selections is detailed below.

\emph{\textbf{QA.}}
Among the QuAC~\cite{Choi:2018:QUA}, CoQA~\cite{Reddy:2019:CQA}, and QReCC~\cite{Anantha:2020:ODQ} datasets, we selected QuAC for QA dialogues. 
QuAC is a widely used dataset for conversational QA and contains 13.6K dialogues between two crowd workers.  
CoQA, on the other hand, is a machine reading comprehension dataset with provided source texts for every dialogue. 
Since these source texts are not necessarily available in real conversations, CoQA was left out. 
QReCC is built based on questions from other datasets, including QuAC and TREC CAsT, and is focused on question rewriting.  Because of the overlapping questions with other datasets, it was also ignored.

\emph{\textbf{Task-oriented.}}
The MultiWOZ~\cite{Zang:2020:MWO} and KVRET~\cite{Eric:2017:KVR} datasets were examined for task-oriented dialogues.
MultiWOZ covers seven various goal-oriented domains: \emph{Attraction, Hospital, Police, Restaurant, Hotel, Taxi, and Train}. 
KVRET, on the other hand, deals with only three domains, all of which are in-car situations.
We, therefore, selected the MultiWOZ dataset, which also has more dialogues than KVRET (8.4K vs. 3K).
Note that MultiWOZ has several versions~\cite{Budzianowski:2018:MWO, Eric:2020:MWO, Zang:2020:MWO}; we used the latest version, MultiWOZ 2.2~\cite{Zang:2020:MWO}.

\emph{\textbf{Social chat.}}
The Wizard of Wikipedia (WoW) ~\cite{Dinan:2018:WWK}, Empathetic Dialogues~\cite{Rashkin:2019:TEO}, Persona-Chat~\cite{Zhang:2018:PDA}, and TaskMaster-1~\cite{Byrne:2019:TTR} datasets were shortlisted for social chat dialogues.
We excluded TaskMaster-1, as the majority of dialogues (7.7K) were collected by crowd workers who were instructed to write full conversations, i.e., played both the user and the system roles on their own.
Persona-Chat and Empathetic Dialogues are more focused on emotional and personal topics, while WoW is knowledge grounded and makes use of knowledge retrieved from Wikipedia.
We therefore chose WoW as a social chat dataset.

Additionally, we also included the TREC 2020 Conversational Assistance Track (CAsT)~\cite{Dalton:2020:CAT} dataset in our study. 
TREC CAsT~\cite{Dalton:2019:CAT}  is an important initiative by the IR community, and is focused on the information seeking aspect of conversations. 
Unlike other datasets, which represent dialogues as a sequence of user-system exchanges, TREC CAsT 2019 provides relevant passages that a system may return in response to a user utterance---therefore, a unique conversation cannot be made for a given conversational trajectory. 
This has been changed in TREC CAsT 2020~\cite{Dalton:2020:CAT}, where a canonical response is given for each user utterance.  We generated conversations for our crowdsourcing experiments using these canonical responses.
In the remainder of this paper we refer to TREC CAsT 2020 as CAsT.

\begin{table}
\caption{Overview of the selected conversational datasets for the entity annotation process. A sample of QuAC, MultiWOZ, and WoW, and all dialogues in TREC CAsT 2020 were used for generating the ConEL dataset.}
\label{tbl:extracted-candidates}
\shrink
\begin{tabular}{@{~}l||@{~~}l|@{~~}l|@{~}l@{~}}
\hline
\textbf{Dataset}	 & \textbf{Task} 				& \textbf{\#Convs~} 	& \textbf{Avg. \#Turns} \\ 
\Xhline{2pt}
QuAC ~\cite{Choi:2018:QUA}& QA & 13.6K 				& 14.5 \\
MultiWOZ ~\cite{Zang:2020:MWO}& Task-oriented 		& 8.4K 					& 13.5 \\
WoW  ~\cite{Dinan:2018:WWK} & Social chat & 22.3K 				& 9.1 \\
TREC CAsT 2020~\cite{Dalton:2020:CAT} & QA & 25 & 17.3 \\
\hline
\end{tabular}
\shrink
\end{table}

\section{Entity Annotation Process}
\label{sec:annot}

\begin{figure}[t]
\fbox{\includegraphics[width=0.9\linewidth]{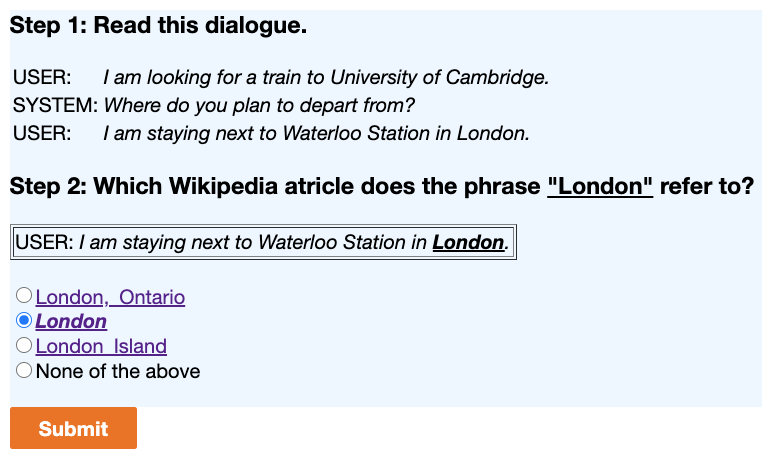}}
  \caption{Annotation interface for the entity-mention selection task (Stage 1). To keep the cognitive load low, only multiple-choice questions were used.  Possible answer entities are linked to their corresponding Wikipedia article.}
 
  \label{fig:stage1_example}

\end{figure}

This section describes the process of annotating dialogues from the selected conversational datasets. 
Our aim is to identify entities that can aid machine understanding of user utterances; this includes named entities, concepts, and mentions of personal entities.
Note that our focus is on user utterances, since the system is supposedly aware of the text it generates during a conversation.  The knowledge graph we use for annotations is Wikipedia (2019-07 dump).

The annotation process was performed via crowdsourcing using Amazon's Mechanical Turk (MTurk). 
In order to reduce the cognitive load for this complex task and to obtain the best annotation results, we ran a number of pilot experiments. 
In these experiments, we tested multiple task structures and interfaces using MTurk and compared the results with expert annotations of the same dialogues. 
The best task designs and interfaces were then used for the final annotations. 
Below, we describe the task design for annotating concepts, named entities, and personal entities (Sections~\ref{sec:annot:con_ens}  and~\ref{sec:annot:pe}), followed by the process of dialogue selection and annotation (Section~\ref{sec:annot:selection}).

%

\subsection{Concepts and Named Entities}
\label{sec:annot:con_ens}

We employ a two-step process for annotating explicitly mentioned entities, i.e., concepts and named entities.

\paragraph{Stage 1: Selecting entity-mention pairs}
First, we aim to map each mention to a single entity in the knowledge graph. 
Workers were presented with a dialogue, a mention from the latest user utterance, and a set of candidate entities or None. 
They were instructed (using a concise description and a couple of examples) to find the Wikipedia article that is referred to by the mention; Figure ~\ref{fig:stage1_example} shows an excerpt from this task. 
The ``None of the above'' option is selected when the candidate pool does not contain the correct entity or the given mention is not appropriate. 
These mentions were later examined by an expert annotator and assigned the correct entity or ignored.
To reduce the cognitive load on the workers, long conversations were trimmed; i.e., the middle turns in conversations with more than six turns were not presented.


\paragraph{Stage 2: Finding the helpful entities}
The mention-entity pairs obtained in the first stage are not necessarily important for machine understanding of user utterances; consider, for instance, the entity \textsc{College} in utterance ``I have wanted to travel to Amsterdam since college. What are the tourist attractions there?''
In Stage 2, we asked workers to filter the entity-mention pairs identified in Stage 1 by selecting only those pairs that can help the system to identify the user's intent. 
Specifically, we provided them with a conversation history and all mention-entity pairs from a user utterance, and gave them the following instruction:
\emph{``Imagine you are an AI agent (e.g., Siri or Google Now), having a dialogue with a person. You have access to Wikipedia articles (and some other information sources) to answer the person's questions. Select the Wikipedia articles that help you to find an answer to the person's question.''}
\fa{We presented mention-entity pairs two times to the users (all in one assignment): once they were asked to select named entities, and the other time they were asked to select all ``helpful'' entities.
Using this interface, we were able to identify named entities and further analyze the differences between concepts and named entities.}


\begin{figure}[t]
\fbox{\includegraphics[width=\linewidth]{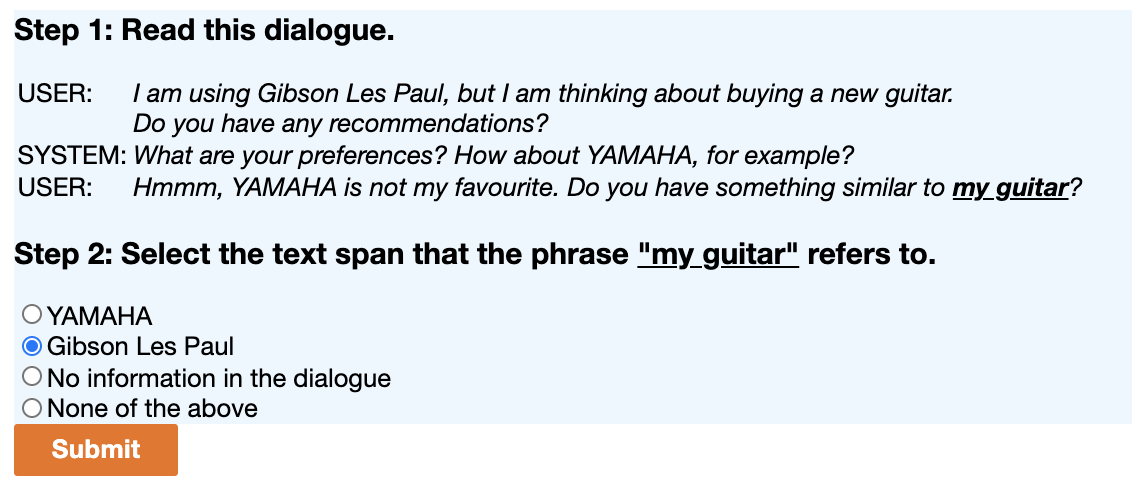}}
  \caption{Annotation interface for mapping a personal entity mention (``my guitar'') to the corresponding explicit entity mention in the conversation (``Gibson Les Paul'').}
 \shrink
  \label{fig:pe_ann_example}
\end{figure}

\paragraph{Generating annotation candidates}
We employ a pooling approach to generate an extended set of candidate mentions and entities.
Three EL tools were used to annotate the dialogues: TagMe~\cite{Ferragina:2010:TAG}, WAT~\cite{Piccinno:2014:FTM}, and REL~\cite{vanHulst:2020:REL}.
Each tool was employed in two ways: (i) the \emph{turn} method, which annotates a single turn, irrespective of the conversation history, and (ii) the \emph{history} method, which annotates each turn given the conversation history up to that turn.  For the CAsT dataset, only user utterances were given to the EL tool, while for other datasets both system and user utterances were considered as conversation history. This  is due to relatively long system utterances in the CAsT dataset, which makes infeasible for the EL tools to annotate the whole conversation history.  
To further improve the recall of our pool, we included the 
top-$10$ Wikipedia search results, using mentions as queries sent to the MediaWiki API.\footnote{\url{https://www.mediawiki.org/wiki/API:Main_page}}


\begin{table*}[t]
\centering
\caption{Entity linking results on the ConEL dataset.} 
\label{tbl:el_all}
\begin{tabular}{l || S rrr | rrr | rrr | rrr || rrr} 
   \hline
	\multirow{2}{*}{\textbf{}}
	& \multicolumn{3}{Sc |}{\textbf{QuAC}} & \multicolumn{3}{c |}{\textbf{MultiWOZ}} 
	& \multicolumn{3}{c |}{\textbf{WoW}} & \multicolumn{3}{c ||}{\textbf{CAsT}} 
	& \multicolumn{3}{c }{\textbf{All}}\\
	&
	\multicolumn{1}{c}{P} & \multicolumn{1}{c}{R} & \multicolumn{1}{c|}{F} & 
	\multicolumn{1}{c}{P} & \multicolumn{1}{c}{R} & \multicolumn{1}{c|}{F} & 
	\multicolumn{1}{c}{P} & \multicolumn{1}{c}{R} & \multicolumn{1}{c|}{F} & 
	\multicolumn{1}{c}{P} & \multicolumn{1}{c}{R} & \multicolumn{1}{c||}{F} & 
	\multicolumn{1}{c}{P} & \multicolumn{1}{c}{R} & \multicolumn{1}{c}{F} \\
   \Xhline{2pt}
	$\textrm{TagMe}_t$ 
	& 29.5	& 37.1	& 32.9
	& 18.5	& 23.9	& 20.9
	& 33.0	& 41.4	& \textbf{36.7}
	& 52.7	& \textbf{47.3}	& 49.8
	& 35.5	& 39.7	& 37.5  \\
	$\textrm{TagMe}_h$
	& 34.7	& 32.4	& 33.5
	& 18.7	& 23.9	& 21.0
	& 33.0	& 41.4	& \textbf{36.7}
	& 57.5	& 46.1	& \textbf{51.2}
	& 38.2	& 38.0	& \textbf{38.1} \\ \hline
	 $\textrm{WAT}_t$
	& 23.2	& 39.0	& 29.1
	& 19.5	& \textbf{36.6}	& \textbf{25.5}
	& 24.8	& 45.7	& 32.2
	& 46.6	& 41.3	& 43.8
	& 28.6	& 40.7	& 33.6 \\
	 $\textrm{WAT}_h$
	& 27.4	& \textbf{48.6}	& 35.1
	& 19.3	& 31.0	& 23.8
	& 23.5	& \textbf{55.7}	& 33.1
	& 44.7	& 45.5	& 45.1
	& 29.6	& \textbf{45.5}	& 35.8 \\ \hline
	 $\textrm{REL}_t$
	& 23.7	& 21.9	& 22.8
	& \textbf{39.3}	& 15.5	& 22.2
	& 43.8	& 10.0	& 16.3
	& 68.1	& 19.2	& 29.9
	& 38.8	& 17.7	& 24.3 \\
	 $\textrm{REL}_h$
	& \textbf{37.6}	& 33.3	& \textbf{35.4}
	& \textbf{39.3}	& 15.5	& 22.2
	& \textbf{52.9}	& 12.9	& 20.7
	& \textbf{70.2}	& 19.8	& 30.8
	& \textbf{47.6}	& 21.3	& 29.4 \\
   \hline
   \end{tabular}
 
 \end{table*} 

\subsection{Personal Entities}
\label{sec:annot:pe}

Annotating conversations with personal entities requires identifying  \emph{personal entity mentions} and mapping them to the corresponding \emph{explicit entity mentions} in the conversation history (if exists); e.g., mapping the personal entity mention ``my guitar'' to the explicit entity mention ``Gibson Les Paul.''
Once this mapping was done, mention-entity pairs can be identified as described in Stage 1 of Section~\ref{sec:annot:con_ens}.
We note that in some cases, explicit entity mentions are not present in the conversation history and the system needs to detect them from other information sources (e.g., previous conversations or user profile data).  
In this study, we confined ourselves to the cases where explicit entity mentions can be found in conversation history; i.e., personal entity mention without explicit entity mention in the conversation history were not mapped to any entity.

We designed a crowdsourcing experiment, where workers were given a conversation history along with the personal entity mention, and their task was to select the text span in the conversation history that the given personal entity mention refers to. 
Figure~\ref{fig:pe_ann_example} shows an example of this task.
``None of the above'' answers were resolved by an expert annotator.



\paragraph{Generating annotation candidates}
We used a simple yet effective method to find personal entity mentions.
Inspired by~\cite{Li:2014:PKG}, we detect all text spans starting with ``my'' followed by one or several adjectives, common nouns, proper nouns and/or numbers (using the SpaCy\footnote{\url{https://spacy.io/}} POS tagger).
We further allowed for the word \emph{``of''} to be part of the mention (e.g., ``my favourite forms of science fiction''). 
For each personal entity mention, we included all the candidate mentions that were identified by our EL methods (cf. Section~\ref{sec:annot:con_ens}).

\subsection{Dialogue Selection and Annotation}
\label{sec:annot:selection}

Annotating all dialogues in the selected datasets was infeasible for us, due to its high costs. 
We therefore selected a random sample of dialogues from a pool of presumably difficult dialogues from the QuAC, MultiWOZ, and WoW datasets. This pool contains dialogues with  at least one \emph{complex mention}, a personal entity mention, or a clarification question in user utterances.
By complex mention we refer to cases where the same mention is linked to different entities by the EL tools (i.e., REL, WAT, and TagMe). 
The clarification questions were identified based on the patterns stated in~\cite{Braslavski:2017:WDY}, and personal entity mentions were extracted as described in Section~\ref{sec:annot:pe}.
A total of $100$ samples were selected ($25$ for each dataset), amounting to $708$ user utterances.
Note that unlike the other datasets, CAsT contains only 25 dialogues, therefore all its dialogues were annotated.
Based on this selection, we are able to analyze the differences between the three main categories of conversational tasks.

%
%
%

The CAsT dataset comes with manually rewritten user queries, where each rewritten query can be answered independently of the conversation history. We also annotated the manually rewritten CAsT queries to allow for a comparison between raw and rewritten queries.
To extend our analysis on personal entity linking, we annotated another sample of dialogues from the WoW dataset. 
To generate this sample, we randomly selected $500$ dialogues that contain personal entity mentions and presented them to crowd workers to find their entity references in the dialogues (cf. Section~\ref{sec:annot:pe}).
Workers agreed that, in 180 dialogues of this sample, the references to the personal entity mentions are present in the dialogue. 
We then randomly selected 25 dialogues (containing 216 user utterances) out of these 180 dialogues and annotated their concepts, named entities, and personal entities. 
%

To ensure high data quality, the annotation tasks were performed by top-rated MTurk workers, i.e., Mechanical Turk Masters.
Since the number of Masters is small, and they mainly select tasks with a high number of HITs, the remaining 0.4\% of our annotation tasks were performed by high quality workers with a task approval rate of 99\% or higher.
We collected three judgments for each annotation and paid the workers \textcent6 for each annotation assignment, resulting in a final cost of around \$620. 
Fleiss' Kappa inter-annotator agreement was 0.76, 0.30, 0.61 for Stage 1, Stage 2, and personal entity annotations, respectively.
Disagreements were resolved by an expert annotator.

%
%


\section{Annotation Results}

In this section, we describe our findings based on the analysis of the entity annotations obtained for the selected datasets. 
We also present baseline results for the entity-annotated conversations.
The results are shown in Tables~\ref{tbl:el_all}--\ref{tbl:el_pe}.
In these tables, the last character of each method, ``t'' or ``h,'' stands for ``turn'' or ``history,'' respectively (cf. Section~\ref{sec:annot:con_ens}).
Precision, recall, and F1 scores are micro-averaged and computed using the strong matching approach~\cite{Usbeck:2015:GGE}. 
To understand the frequency of personal entities in conversational datasets, we applied the method described in Section~\ref{sec:annot:pe} to identify all personal entity mentions in all the datasets.
We found that WoW contains more dialogues with personal entity mentions compared to other datasets; i.e., 33\% of dialogues in WoW vs. 0.3\%, 11\%, and 12\% of dialogues in QuAC, MultiWOZ, and TREC CAsT, respectively.
These results indicate that \textbf{personal entity mentions are mainly present in social chat conversations}.

Comparing concepts and named entities, we found that 43\% of linked entities in the ConEL dataset are marked as named entities by crowd workers, which implies that the remaining 57\% entities are concepts.
This indicates that \textbf{in addition to named entities, concepts are also found useful for understanding the intents of user utterances.} 

%


Table~\ref{tbl:el_all} shows the results of different EL methods on the ConEL dataset. While TagMe achieves the highest F1 scores on WoW and CAsT, WAT and REL are the best performing tools (with respect to F1) on the MultiWOZ and QuAC datasets, respectively.
Comparing the ``turn'' and ``history'' methods, we observe that conversation history improves EL results for most datasets and tools. 
We also find that REL has higher precision but lower recall compared to TagMe and WAT. 
One might argue that high precision EL is preferred in a conversational setting, as incorrect results can lead to high user dissatisfaction. This claim, however, requires further investigation, and the effect of EL on end-to-end conversational system performance is yet to be evaluated.


Table~\ref{tbl:el_breaksown} compares EL results for named entities and concepts separately, where F1 scores are computed based on only named entities $\textrm{F}_{\textrm{NE}}$ or concepts $\textrm{F}_{\textrm{C}}$.
We observe that REL is better at linking named entities, while TagMe is better at linking concepts. 
This shows that although it is important to achieve high performance for both named entities and concepts, there is no single EL tool that excels at both. 
The results in Tables~\ref{tbl:el_breaksown} and~\ref{tbl:el_all} suggest that \emph{\textbf{all existing EL tools that we examined are suboptimal for EL in a conversational setting.}}

\begin{table}[t]
\centering
\caption{Breakdown of entity linking results for named entities ($\textrm{F}_{\textrm{NE}}$) and concepts ($\textrm{F}_{\textrm{C}}$).} 
\label{tbl:el_breaksown}

\begin{tabular}{@{~}l||@{~}Sr@{~~}r|@{~}r@{~~}r|@{~}r@{~~}r|@{~}r@{~~}r||@{~}r@{~~}r@{~}}
	\hline
	\multirow{3}{*}{}
	& \multicolumn{2}{@{~}Sc|@{~}}{\textbf{QuAC}} & \multicolumn{2}{@{~}c|@{~}}{\textbf{Multi}} 
	& \multicolumn{2}{@{~}c |@{~}}{\textbf{WoW}} & \multicolumn{2}{@{~}c ||@{~}}{\textbf{CAsT}} 
	& \multicolumn{2}{@{~}c }{\textbf{All}} \\
	& &  & \multicolumn{2}{c |@{~}}{\textbf{WOZ}} & && && & \\
	& 
	$\textrm{F}_{\textrm{NE}}$ & $\textrm{F}_{\textrm{C}}$ &
	$\textrm{F}_{\textrm{NE}}$ & $\textrm{F}_{\textrm{C}}$ &
	$\textrm{F}_{\textrm{NE}}$ & $\textrm{F}_{\textrm{C}}$ &
	$\textrm{F}_{\textrm{NE}}$ & $\textrm{F}_{\textrm{C}}$ &
	$\textrm{F}_{\textrm{NE}}$ & $\textrm{F}_{\textrm{C}}$ \\
	\Xhline{2pt}
	$\textrm{TagMe}_t$
	& 32.2	& 6.3	& 17.3	& 9.4	& \textbf{34.6}	& 11.8	& 24.4	& 40.7	& 27.5	& 21.6 \\
	$\textrm{TagMe}_h$
	& 30.5	& \textbf{11.3}	& 15.9	& 11.0	& \textbf{34.6}	& 11.8	& 24.3	& \textbf{43.3}	& 26.5	& \textbf{24.4} \\\hline
	$\textrm{WAT}_t$
	& 25.0	& 8.9	& 15.5	& \textbf{15.4}	& 23.8	& 15.0	& 22.6	& 35.1	& 22.1	& 20.2 \\
	$\textrm{WAT}_h$
	& 31.7	& 8.5	& 14.8	& 14.7	& 22.4	& \textbf{16.2}	& 22.1	& 35.9	& 23.9	& 20.7 \\\hline
	$\textrm{REL}_t$
	& 26.1	& 0.0	& \textbf{31.7}	& 3.1	& 18.2	& 8.5	& 63.8	& 2.4	& 35.1	& 2.5 \\
	$\textrm{REL}_h$
	& \textbf{40.7}	& 0.0	& \textbf{31.7}	& 3.1	& 25.0	& 8.3	& \textbf{66.0}	& 2.4	& \textbf{43.1}	& 2.5  \\
\hline
\end{tabular}
\shrink
\end{table}


Table~\ref{tbl:el_cast} shows the EL results for all raw and re-written CAsT queries.
Similar to Table~\ref{tbl:el_all}, we observe that there is a trade-off between precision and recall across the different EL tools. 
The results also show higher scores for rewritten queries compared to raw queries, which is due to resolved coreferences and richer context in the rewritten queries.


Table~\ref{tbl:el_pe} shows EL results on a sample of WoW dialogues, all containing references to personal entities (cf. Section~\ref{sec:annot}). This sample is annotated with concepts, named entities, and personal entities.
The left block shows the results of different EL methods in their original form, i.e., without annotating personal entity mentions.
The right block in Table~\ref{tbl:el_pe} represents a modified version of the same methods, where each method is extended to identify and link personal entity mentions, denoted with \emph{PE}. 
Considering a personal entity mention $m_{pe}$, and an entity $e$, the PE method computes the cosine similarity between the word embedding of $m_{pe}$ and the entity embedding of entity $e$. 
For every $m_{pe}$, we compute this similarity with all the previously linked entities in the conversation and find the most similar entity.
Mention-entity pairs $\langle m_{pe}, e \rangle$ below a certain threshold $\tau$ are ignored. This threshold allows for filtering personal entity mentions that do not have the corresponding entities in the conversation history.
We used Wikipedia2Vec~\cite{Yamada:2020:W2V} word and entity embeddings released by \citet{Gerritse:2020:GEE}.
The threshold $\tau$ was set empirically by performing a sweep (on the range [0, 1] in steps of 0.1) using 5-fold cross-validation. 

Comparing the left and right parts of Table~\ref{tbl:el_pe}, we observe a slight (albeit often negligible) performance increase for the PE method. These results show that identifying personal entity mentions and their corresponding entities is a non-trivial task and cannot be resolved with a simple extension of current approaches. This reinforces our finding that all examined EL tools are suboptimal for EL in conversations.

\begin{table}[t]
\centering
\caption{Entity linking results on TREC CAsT raw and re-written dialogues.} 
\label{tbl:el_cast}
\begin{tabular}{l|| S ccc | ccc } 
   \hline
	\multirow{2}{*}{\textbf{}}
	& \multicolumn{3}{Sc |}{\textbf{CAsT (raw)}} & \multicolumn{3}{c }{\textbf{CAsT (re-written)}} \\
	&
	P & R & F & 
	P & R & F \\
   \Xhline{2pt}
	 $\textrm{TagMe}_t$
	& 52.7	& \textbf{47.3}	& 49.8
	& 64.1	& \textbf{66.4}	& \textbf{65.2}  \\
	 $\textrm{TagMe}_h$
	& 57.5	& 46.1	& \textbf{51.2}
	& 65.6	& 64.6	& 65.1 \\ \hline
	 $\textrm{WAT}_t$
	& 46.6	& 41.3	& 43.8
	& 52.9	& 45.3	& 48.8 \\
	 $\textrm{WAT}_h$
	& 44.7	& 45.5	& 45.1
	& 54.9	& 50.8	& 52.8 \\ \hline
	 $\textrm{REL}_t$
	& 68.1	& 19.2	& 29.9
	& 73.8	& 27.1	& 39.6 \\
	 $\textrm{REL}_h$
	& \textbf{70.2}	& 19.8	& 30.8
	& \textbf{76.4}	& 27.9	& 40.8 \\
   \hline
   \end{tabular}
 
 \end{table}


\begin{table}[t]
\centering
\caption{Entity linking results on a sample of the WoW collection containing references to personal entities. The left block shows the results of the original EL methods and the right block represents the results a of modified EL methods, where personal entities are also annotated.} 
\label{tbl:el_pe}

\begin{tabular}{l ||@{~~}S l@{~~}S l@{~~}S l l } 
    \hline
	& 
	P & R & F \\
   \Xhline{2pt}
	$\textrm{TagMe}_t$ & 62.2	&	50.2	&	55.6 \\
	$\textrm{TagMe}_h$ & 62.4	&	49.6	&	55.3 \\ \hline
	$\textrm{WAT}_t$  & 56.0	&	63.1	&	59.4\\
	$\textrm{WAT}_h$  & 55.6	&	\textbf{67.1}	&	\textbf{60.8}\\ \hline
	$\textrm{REL}_t$ & 64.0	&	18.0	&	28.1\\
	$\textrm{REL}_h$  & \textbf{78.0}	&	22.0	&	34.3\\
   \hline
\end{tabular}
\, 
\begin{tabular}{l ||@{~~}S l@{~~}S l@{~~}S l l } 
   \hline
	& 
	P & R & F \\
   \Xhline{2pt}
	$\textrm{TagMe}_t$+PE &	61.7	&	51.1	&	55.9 \\
	$\textrm{TagMe}_h$+PE  &	62.0	&	50.7	&	55.8 \\ \hline
	$\textrm{WAT}_t$+PE &	56.2	&	64.4	&	60.0	\\
	$\textrm{WAT}_h$+PE &	55.7	&	\textbf{68.8}	&	\textbf{61.6}	 \\ \hline
	$\textrm{REL}_t$+PE &	63.2	&	18.6	&	28.8	\\
	$\textrm{REL}_h$+PE &	\textbf{77.2}	&	22.6	&	34.9	\\
   \hline
\end{tabular}
\end{table}

\section{Conclusion}
In this paper, we studied entity linking in a broad setting of conversational systems: QA, task-oriented, and social chat.
Using crowdsourcing, we analyzed existing conversational datasets and annotated them with concepts, named entities, and personal entities.
We found that while both concepts and named entities are useful for understanding the intent of user utterances, personal entities are mainly important in social chats.
Further, we compared the performance of different established EL methods in a conversational setting and concluded that none of the examined methods can handle this problem effectively, falling short in providing both high recall and precision, as well as annotating concepts, named entities, and personal entity mentions. Our annotated conversational dataset (ConEL) and interface designs are made publicly available. These resources come with detailed instructions on the procedure of collecting the annotations, which can be used for further extension of the collection. 
Following the insights from this study, developing conversational entity linking methods and employing them in various types of conversational systems are obvious future directions.

\balance
\bibliographystyle{ACM-Reference-Format}
\bibliography{references-base}


\end{document}
\endinput